\documentclass[a4paper, 9pt, conference]{ieeeconf}      

\IEEEoverridecommandlockouts                              

\usepackage{lipsum}
\usepackage{graphicx}                       
\usepackage{stfloats}
\usepackage[tight,footnotesize]{subfigure}  

\usepackage{amssymb,amsmath}
\usepackage{textcomp}
\usepackage{mdwmath}
\usepackage{mdwtab}
\usepackage{eqparbox}
\usepackage{rotating}                       
\usepackage{array}                          
\usepackage{listings}                       
\usepackage[ruled,vlined,noresetcount]{algorithm2e}
\usepackage{listings}
\usepackage[noend]{algpseudocode}
\usepackage{bm}
\usepackage[colorlinks,
            linkcolor=blue,
            anchorcolor=blue,
            citecolor=blue,
            bookmarks=false
            ]{hyperref}
\usepackage{cite}

\usepackage{amsmath}
\usepackage{amssymb}
\usepackage{eucal}
\usepackage{graphicx}
\usepackage{float}
\usepackage{subfigure}
\usepackage{amsmath}
\usepackage{algpseudocode}
\usepackage{stfloats}
\usepackage{multirow} 
\usepackage{xcolor}
\usepackage[ruled,vlined]{algorithm2e}
\usepackage{amssymb}
\usepackage{booktabs}

\title{\LARGE \bf Semi-Supervised Confidence-Level-based Contrastive Discrimination for  Class-Imbalanced Semantic Segmentation}

\author{Kangcheng Liu 
\thanks{$^{1}$K. Liu is with the Department of Mechanical and Automation Engineering, The Chinese University of Hong Kong, Shatin, N.T.,  Hong Kong 999077, China.
(email: {\tt\small kcliu@mae.cuhk.edu.hk}).  K. Liu is the corresponding author.}
}

\begin{document}

\maketitle
\thispagestyle{empty}
\pagestyle{empty}

\begin{abstract}
 To overcome the data hungry challenge, we have proposed a semi-supervised contrastive learning framework for the task of class-imbalanced semantic segmentation. First and foremost, to make the model operate in a semi-supervised manner, we proposed the confidence-level-based contrastive learning to achieve instance discrimination in an explicit manner, and make the low-confidence low-quality features align with the high-confidence counterparts. Moreover, to tackle the problem of class imbalance in crack segmentation and road components extraction, we proposed the data imbalance loss to replace the traditional cross entropy loss in pixel-level semantic segmentation. Finally, we have also proposed an effective multi-stage fusion network architecture to improve semantic segmentation performance. Extensive experiments on the real industrial crack segmentation and the road segmentation demonstrate the superior effectiveness of the proposed framework. Our proposed method can provide satisfactory segmentation results with even merely 3.5\% labeled data. 
\end{abstract}

\section{Introduction}
\begin{figure*}[t!]
\centering
\includegraphics[scale=0.386]{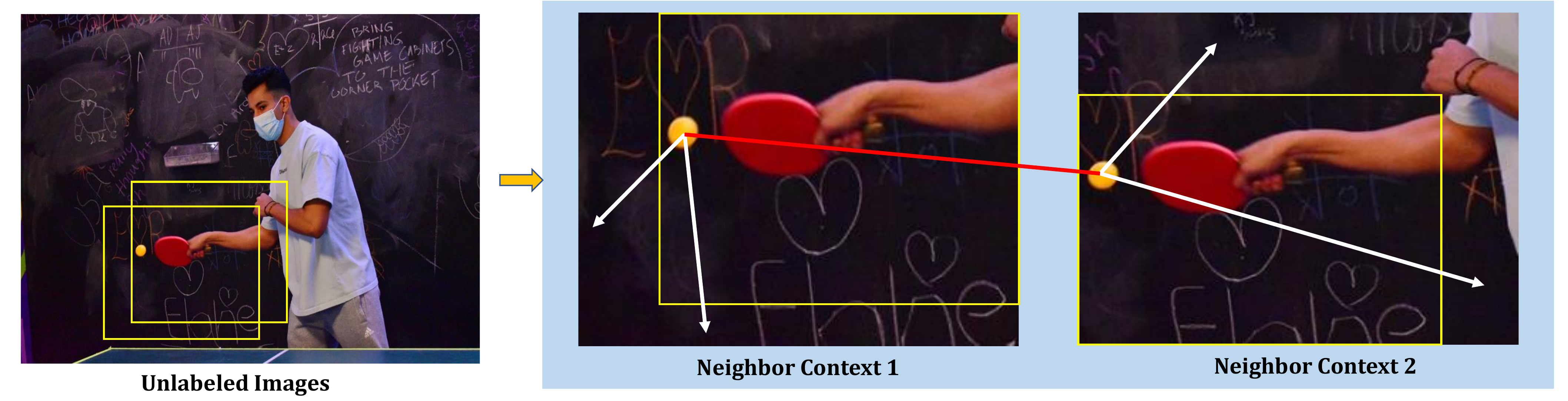}
\caption{The neighbour context one and neighbour context two are randomly cropped from the same image with the overlapping regions. We keep the consistency (denoted by the red line) within the diverse contexts (denoted by the white line.) in pixel level.}
\label{fig_context}
\end{figure*}

Currently, the semantic segmentation methods has achieved great success throughout the past years both in 2D images \cite{liu2019deep} and 3D point clouds \cite{liu2022fg, liu2022weaklabel3d, liu2022weakly, liuws3d, liu2022robustLiDAR,liu2022robust}. However, the data hunger problem has longly confused the research community. And it can be very labor-intensive and ineffective for human annotators to make pixel-level annotations. For instance, it takes approximated 1.8h for the annotation of a single image in the typical semantic segmentation dataset. And the heavy reliance of current deep learning-based semantic segmentation approaches on abundant richly labeled training data has prevented the application of them into real industrial applications, such as the infrastructural crack segmentation, and the industrial road segmentation in the transportation networks. Typically, the LiDAR is still the most popular and prominent sensor because it is the one of the most fast and accurate robotic sensors \cite{liu2017avoiding, liu2020fg, liu2022weaklabel3d, liu2021fg, liu2019deep, liu2022fg, zhao2021legacy, liu2022ARM, liu2022CYBER2, liu2022ICCA1, liu2022ICCA2}. However, the visual systems such as the cameras are also important for perceiving and interacting with the real-world. 
To reduce the reliance on large amount of high-quality labels, various forms of annotations have been proposed, such as the detection box-level annotation \cite{liu2022weaklabel3d}, the image class level-label \cite{jing2019coarse}, and the scribble-based label \cite{yuzhi2020legacy}. In this work, we concentrate on the semi-supervised semantic segmentation which means there are merely a small portion of labeled data and much greater percentage of unlabeled data. 
\begin{figure}[t!]
\centering
\includegraphics[scale=0.362]{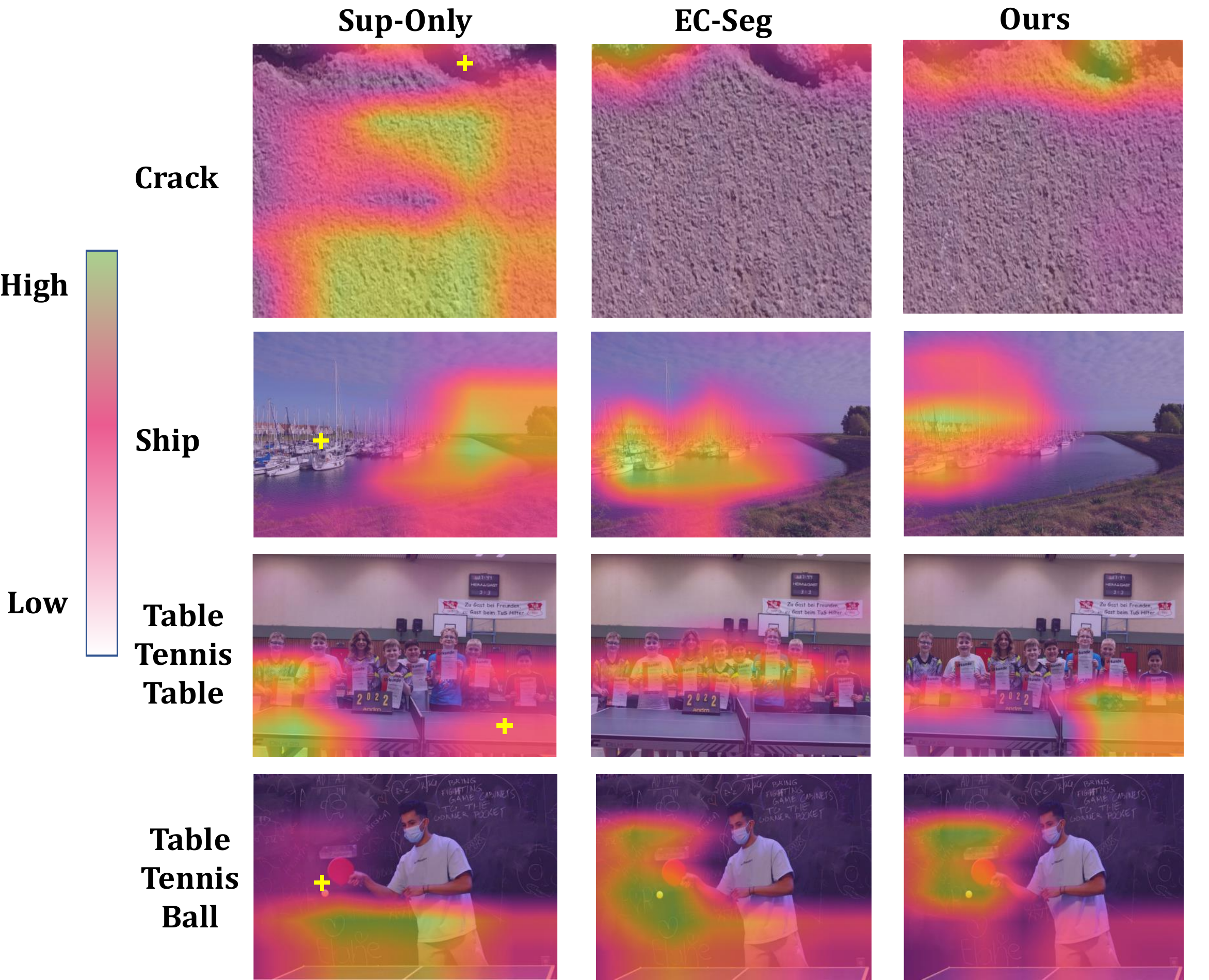}
\caption{The gradient-based Class Activation Map visualization of the neighbouring contextual contributions to the query interested pixel (the golden yellow cross). The activation bar is shown in the left of the figure. The sup-only model is trrained with merely 3.5\% labeled training data. We have compared our proposed approach with the State-of-the-art EC-Seg \cite{mendel2020semi}.}
\label{fig_cam}
\vspace{-3mm}
\end{figure}
It has also been demonstrated by various of mature frameworks that the contextual information plays a very significant role in semantic segmentation. The mature frameworks such as the ASPP \cite{yang2018denseaspp} and CE-Net \cite{zhang2018context} have shown great performance by considering the semantic correlations in the neighbouring pixels. In the semi-supervised learning circumstances, the model can easily overfit the emerged semantic classes in the very limited training data. And when encountering the novel classes, the model tends to have poor generalization capacity, especially in the class imbalanced circumstances. We have visualized in Fig. \ref{fig_cam}. It can be seen that we merely 3.5\% of images in the training set are labeled for the smi-supervised setting, the model tends to overlook the contextual information and pay no attention to the correlated semantic class, which means the contextual relationship is lost. 
In this work, we have proposed a simple but effective contextual contrastive learning framework for semi-supervised semantic segmentation. As shown in Fig. \ref{fig_context}, the images are cropped with a overlapped region, and the same pixel are put within diverse contexts to achieve context awareness. Also, we propose the directional contrastive learning which encourages the pixel-level feature to converge towards a highly confident prediction, which provided a more reliable results. Also, to tackle the class imbalance problem in our industrial applications of road and crack semantic segmentation, we have proposed the data imbalance loss to substitute the original cross-entropy loss in the task of pixel-level semantic segmentation. The CAM-based semantic segmentation activation is shown in Fig. \ref{fig_cam}. It can be demonstrated that our proposed method can capture better contextual relations and can make the contextual information participate in the prediction of the semantic classes of the query pixel. And the activation in the pixels that are of the same class as the query pixel is significantly larger.

The proposed framework can achieved State-of-the-art (SOTA) semantic segmentation performance in the semi-supervised setting, and can be integrated seamlessly to current state-of-the-art (SOTAs) semantic segmentation frameworks with a few added computational costs. Extensive experiments on the real-site road semantic segmentation and the crack semantic segmentation demonstrate the superior effectiveness of our proposed approaches.
In summary, the prominent contributions of this work is:
\begin{enumerate}
\item To achieve context awareness, we propose using context-aware pixel-level contrastive learning to make our model achieve prediction consistency under various of transformations.
\item To achieve context consistency, we propose the first confidence-level based contrastive loss, which encourage the high-confident network predictions.
\item  To eliminating overfitting in the class-imbalanced segmentation, we propose the data balance loss to replace the traditional cross entropy loss. 

\item A backbone network \textbf{CR-Seg} is also proposed. Extensive experiments on class-imbalanced crack segmentation and road segmentation demonstrates the effectiveness of the proposed framework.

\end{enumerate}

 \section{Methodology} 
 We have proposed the first framework of confidence-level-based contrastive learning for the class-imbalanced semantic segmentation. The network framework consistsof the labeled branch and the unlabeled branch to achieve the semi-supervised semantic segmentation. Our proposed framework is consist of three parts: 1): Firstly, we proposed the context-aware contrastive learning and the confidence-level-based contrastive loss, which is illustrated in Section \ref{sec_confidence}.  2): We propose the data balance loss to replace the traditional cross-entropy loss as well as the other optimization functions to regularize the labeled data, which is detailed in Section \ref{sec_data_im}. 3): We also propose the network frameworks to tackle the class-imbalanced crack semantic segmentation and road segmentation, respectively, which is detailed in Section \ref{sec_net_arc}.
 \begin{figure*}[t!]
\centering
\includegraphics[scale=0.256]{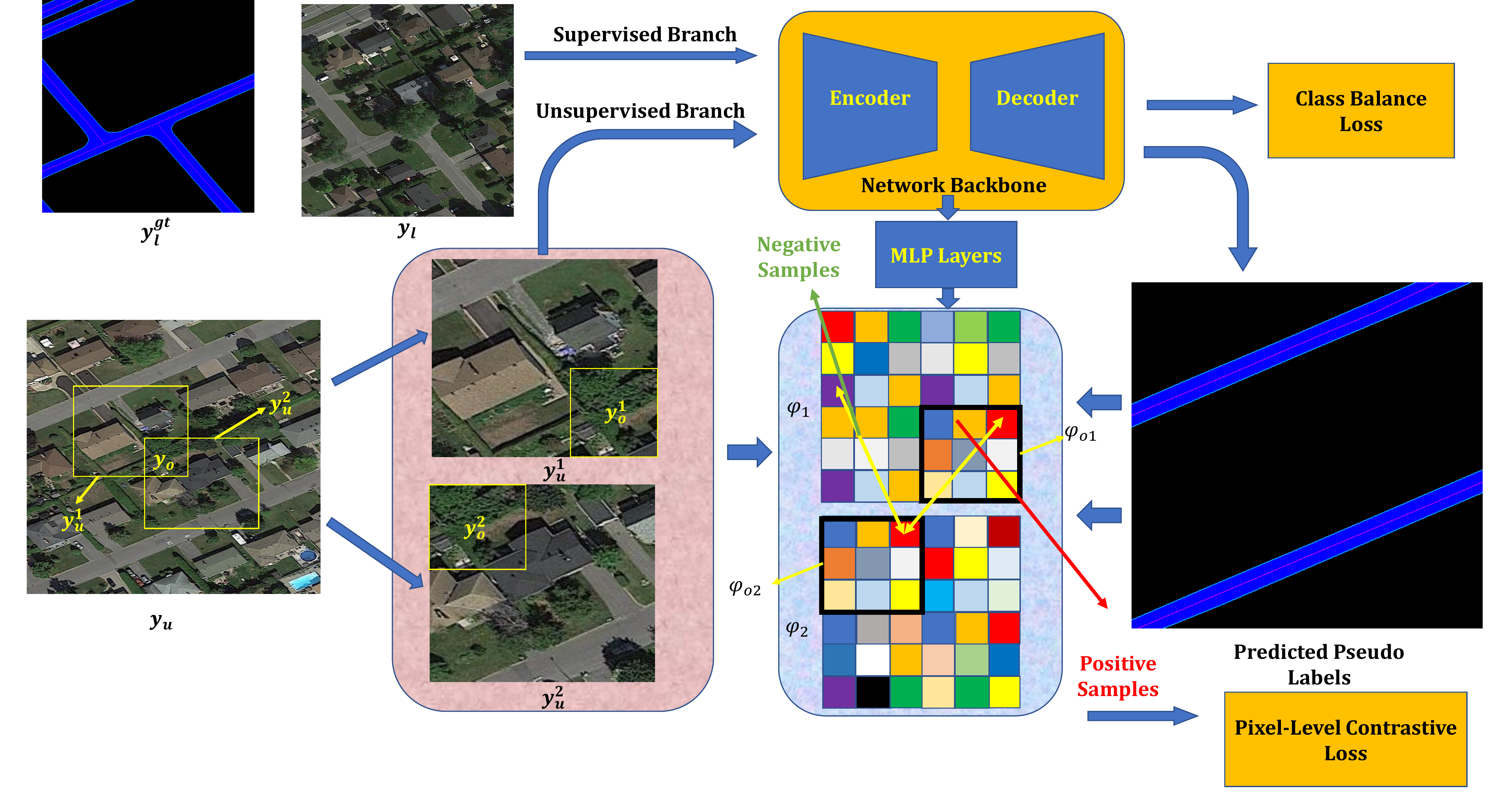}
\caption{Our proposed framework consists of the labeled branches, which consumes the labeled data, and the unlabeled branches, which consumes the unlabeled data. In the unlabeled branch, we encourage the semantic prediction consistency between the same image patch under two diverse contexts.}
\label{fig_framework}
\vspace{-4mm}
\end{figure*}
 \subsection{Confidence-Level based Contrastive Learning for semi-supervised learning}
 \label{sec_confidence}
 We have shown our system framework in Fig. \ref{fig_framework}.
 In the semi-supervised setting, we can have the labeled data $y_l$ and the unlabeled data $y_u$. For the labeled data $y_l$, like the traditional semantic segmentation model, we can utilize them to pass through the network $\varphi$ to obtain the feature representations. Finally, the semantic prediction is regularized by our proposed optimization functions in the Subsection \ref{sec_data_im}, which is similar to our previous work \cite{liu2019deep}. 
 For the unlabeled data $y_u$, two random patches $y^1_u$ and $y^2_u$ are cropped. Utilizing the multi-layer perceptron (MLP), the two patches can be transformed to the enhanced feature representations $\varphi_1$ and $\varphi_2$. The features in the overlapped regions are denoted as $y^1_o$ and $y^2_o$, respectively. We proposed contextual consistent semantic contrastive optimization function, which encourages the overlapped region $y_o$ to be consistent under various background contexts, which are corresponding to $y^1_u$ and $y^2_u$, respectively.

 As shown in Fig. \ref{fig_framework}, we propose our confidence-level-based contrastive loss function that operates at the pixel level. To conduct effective constrastive learning, firstly, we should select appropriate positive and negative pairs. We regard the same pixels (i.e. the same pixels that lie in $\varphi_{o1}$ and $\varphi_{o2}$) under different two different contexts mentioned above as the positive pairs, because they definitely have the same semantic classes. And we regard any different pixel pairs in $\varphi_{1}$ and $\varphi_{2}$, as the negative pairs.
 After determining the positive and negative pairs, we can formulate the confidence-level based contrastive loss functions as follows:
 
\begin{tiny}
\begin{equation}
L^{m}_{contrast}=-\frac{1}{N_p}\sum_{(o1, o2) \in S_p} log\frac{H_{o1, o2}exp(\varphi_{o1}^{+} \cdot \varphi_{o2}^{+}/\tau)}{\sum_{(o1, o2^{-}) \in S_p}  exp(\varphi_{o1}^{+} \cdot \varphi_{o2}^{-}/\tau))},
\end{equation}
\end{tiny}
Denote the final maximum probabilistic predictions among all $c$ classes as $\mathcal{P}^{c}_{o1}$ and $\mathcal{P}^{c}_{o2}$ respectively, we denotes the confidence indicator $H_{o1, o2}$ as follows: 
\begin{equation}
    H_{o1, o2}= \textbf{1} \{\,  \alpha < \mathcal{P}^{c}_{o1} < \mathcal{P}^{c}_{o2}\ \}
\end{equation}
Therefore, all the probabilistic predictions are made to converge to the ones with higher confidence value. And with the training of the network, merely the highly confident predictions are retained. And the highly confident predictions are prevented from being mistakenly guided by the less confident predictions. Also, a threshold $\alpha$ is proposed to filter out those samples that are not very confident. Therefore, by the proposed confidence-level based contrastive discrimination, we have not only filtered out those samples that are not confident enough, but also make alignment of those pixel-level features to their highly confident counterparts. The above two improvements can both contribute to the final semantic segmentation performance. 
\subsection{Other Optimization Function Formulations for the Supervised Branch}
 \label{sec_data_im}
 We have also proposed the data balance loss and the construction loss for the labeled data in the supervised branch. The weight decay loss is both utilized for the labeled data in the supervised branch and unlabeled data in the unsupervised branch.
 \subsubsection{Data Balance Loss Function}
 We have also proposed the data balance loss to tackle the class imbalance problem for the limited labeled data. Because in our applications of crack semantic segmentation and road semantic segmentation, the problems can be formulated as a extremely class imbalanced two-class semantic segmentation problem. Denote the total number of pixels in class one and class two as $N_1$ and $N_2$ respectively, the data balance loss can be formulated as follows:
\begin{equation}
\scriptsize
        L^{m}_{Balance}=-\frac{1}{N_i}\sum_{i=1}^{N_i} {\lambda_1(1-b^{gt}_i)}^{\alpha} log(b_i)+{\lambda_2(b^{gt}_i)^{\alpha}} log(1-b_i),
\vspace{-1.6mm}
\end{equation}
The value of $\lambda_1$ and $\lambda_2$ are given as: $\lambda_1=\frac{N_1}{N_{total}}$ and $\lambda_2=\frac{N_2}{N_{total}}$. 

In our applications of crack and road semantic segmentation, the class one can be the crack pixels and the road pixels, and the class two can be the non-crack pixels and the non-road pixels.
 



\subsubsection{The construction loss function}
For the supervised branch, although the $L_{Balance}$ loss mentioned above provides pretty good fitting between the data distribution and the trained discriminative distribution, there are still overfitting problems, especially in extremely imbalanced class distribution cases.
The construction loss is proposed to solve such issue in the training phase, which is formulated as:
\begin{equation}
    L_{c}=||P-y||_2^2
\end{equation}

where $||\cdot||$ refers to the $\mathcal{L}$-2 norm, which is also known as least squares. $L_{c}$ is basically minimizing the sum of square of the differences between the predicted map and ground truth. 
\begin{figure}[t]
\centering
\includegraphics[scale=0.31]{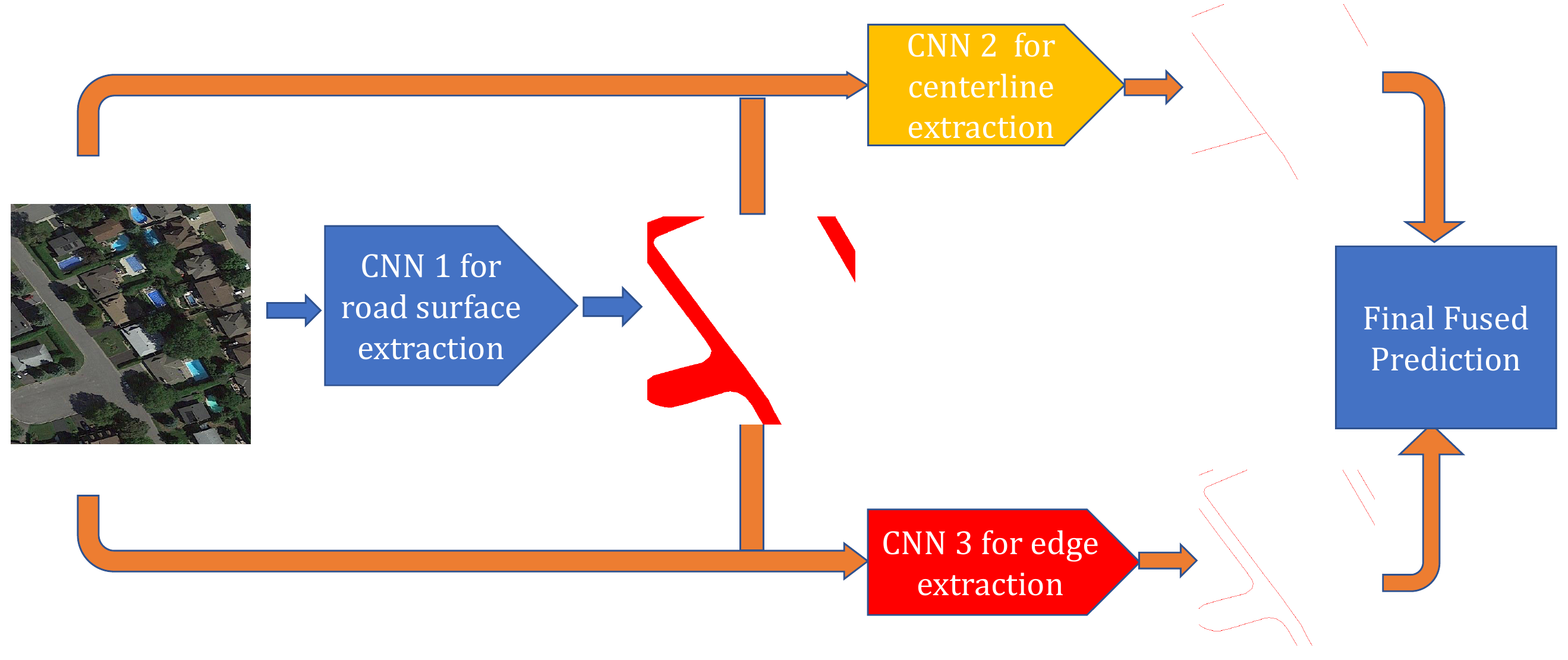}
\caption{The road extraction network}
\label{fig_cnn_road}
\end{figure}

\begin{figure}[h!]
\centering
\includegraphics[scale=0.25]{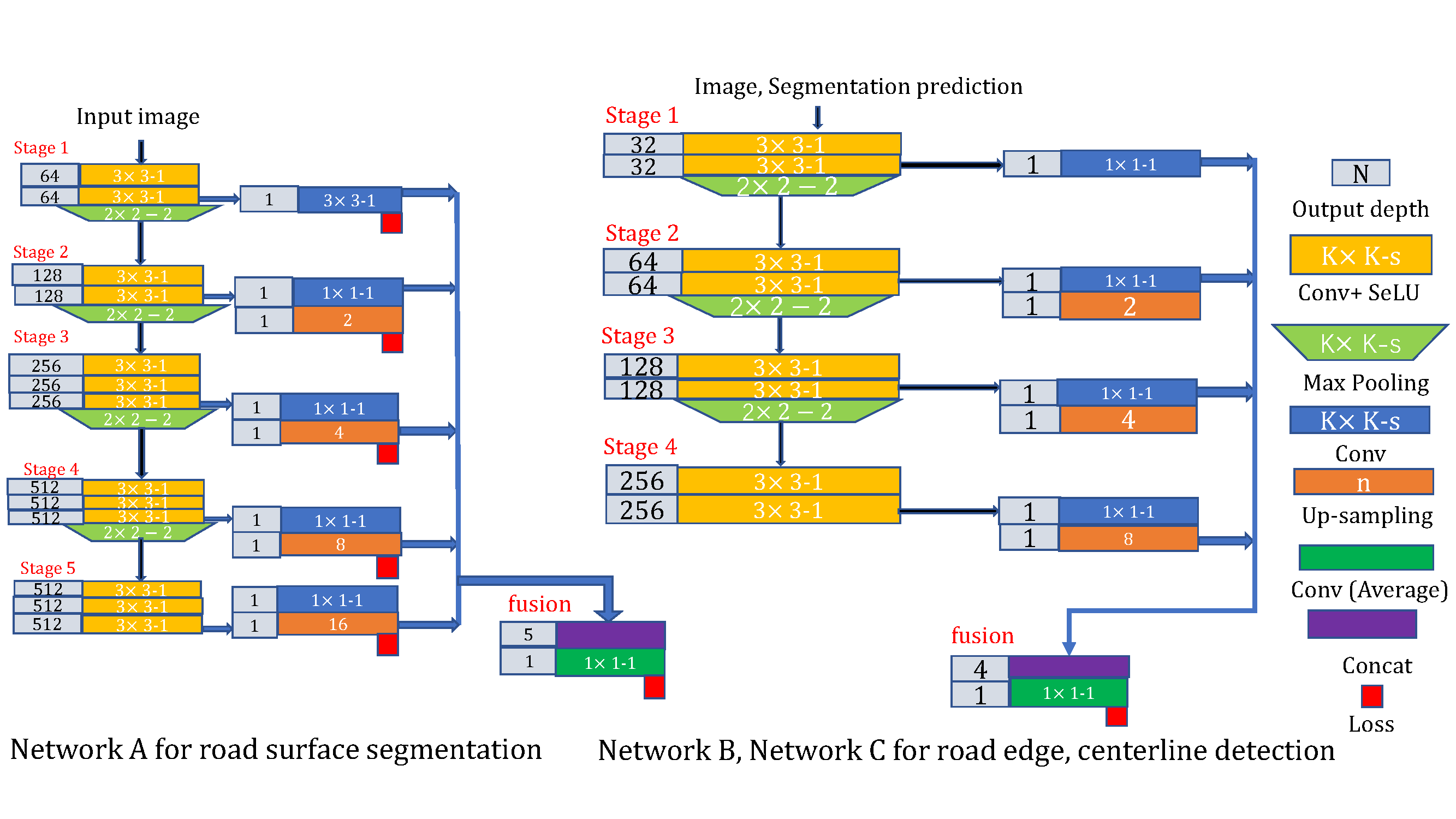}
\caption{The proposed multi-stage-fusion crack/road semantic segmentation network structure}
\label{fig_roadnet}
\end{figure}

\subsubsection{Weight decay loss}
Weight decay is commonly adopted regularization method for the optimization of the model parameters. It can suppress all the irrelevant components of the weight vector by choosing the smallest vector that solves the learning task, which is denoted as:
\begin{equation}
L_w=\frac{\lambda}{2}||W||_2^2
\end{equation}
where $\lambda$ is a hyper-parameter governing how strongly large weights are penalized and it is set to 
a constant value: 2e-4.
\subsubsection{The overall loss function}
Finally the overall loss function can be formulated as the sum of the proposed losses and the fusion of losses at multiple stages. Denote the number of stages of the network as $M$, the total loss $L_{multi}$ is as follows:
\begin{equation}
L_{multi}=L_{c} + L_{w} + \sum_{m=1}^{M} L_{contrast}^m + L^{m}_{Balance} , 
\label{multi}
\end{equation}




 
\begin{figure}[t!]
    \centering
    \includegraphics[scale=0.25]{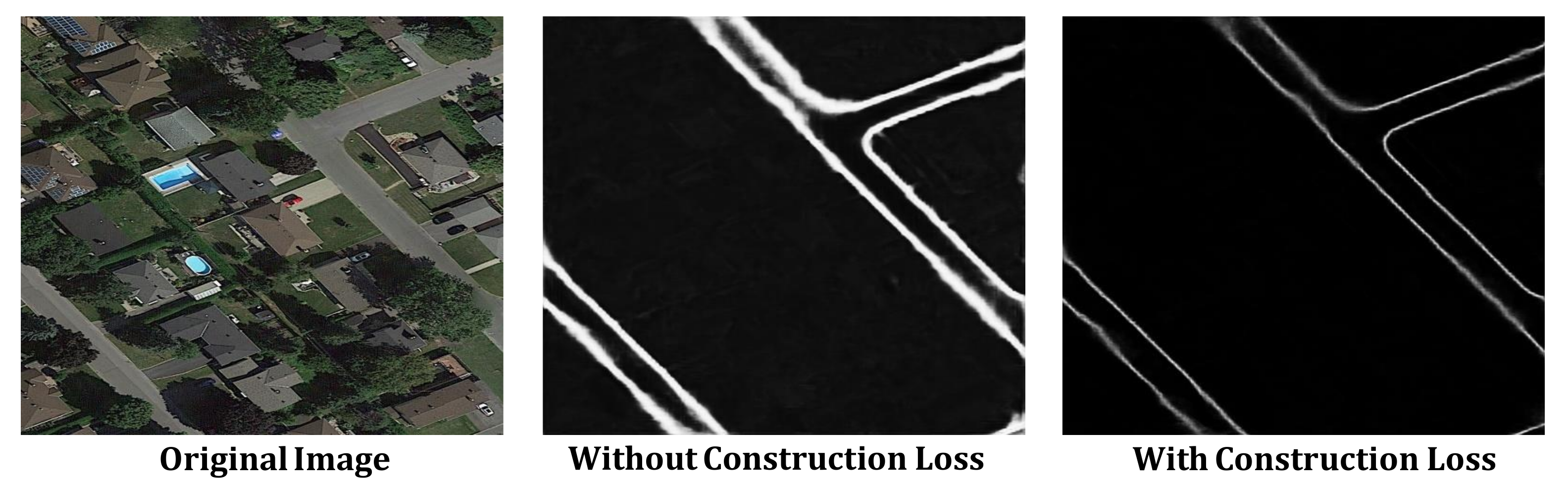}
    \caption{The comparison of segmentation result with and without contrastive loss for the labeled data.}
    \label{fig_cons_loss}
\end{figure}

\begin{figure}[t!]
    \centering
    \includegraphics[scale=0.39]{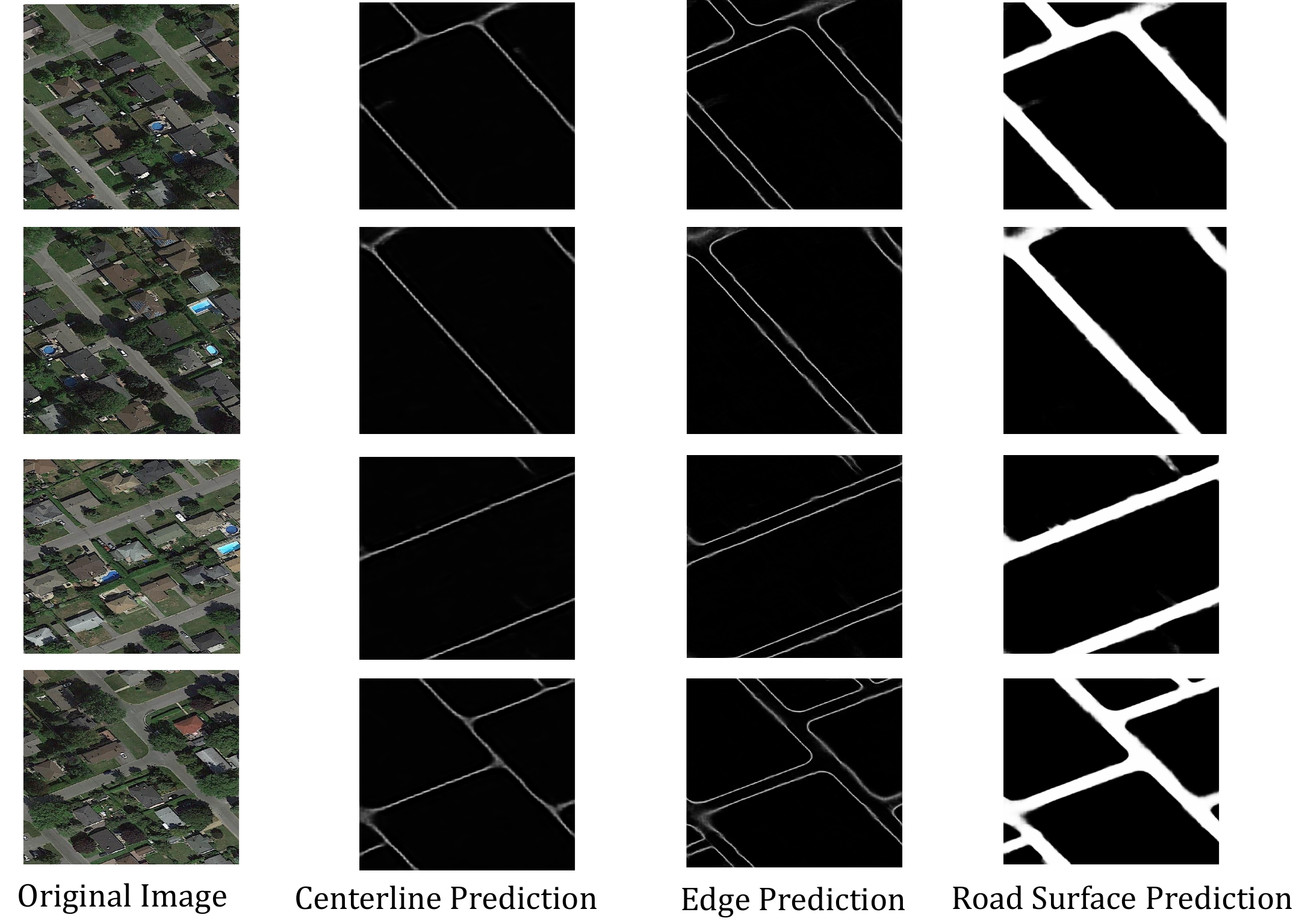}
    \caption{The road centerline, edge, surface segmentation result with merely 3.5\% labeled images.}
    \label{edge_results}
    \vspace{-3mm}
\end{figure}


\begin{figure}[t!]
    \centering
    \includegraphics[scale=0.25]{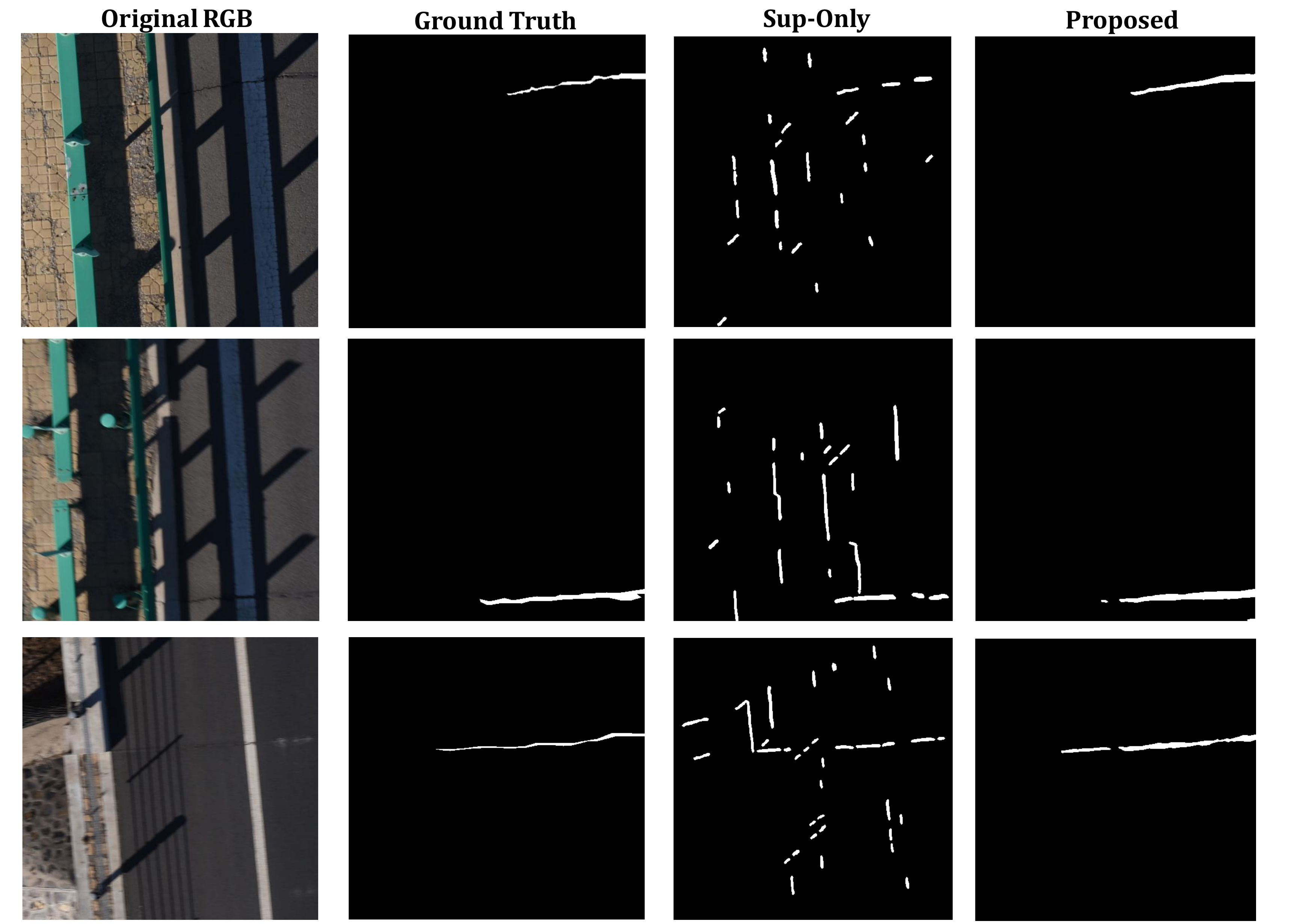}
    \caption{The comparison of our method with the Sup-Only method, for the task of crack semantic segmentation with merely 3.5\% labeled data.}
    \label{semi_crack}
\end{figure}
\subsection{Semantic Segmentation Network Architecture}
\label{sec_net_arc}
\subsubsection{Network Selections for the Crack Segmentation and the Road Extraction}
In this Section, we have proposed network architectures \textbf{CR-Seg} that can be both used for crack semantic segmentation and road semantic segmentation, which is an improved architecture based on our previous work \cite{liu2019deep}. It is  For the crack semantic segmentation, we just use a single segmentation network to complete this task.
We have used three convolutional Neural Networks which are utilized for road surface segmentation, road edge detection and road center-line extraction, which is shown in Figure \ref{fig_cnn_road}. The road surface prediction is utilized as the input to the road centerline and road edge detection because the road surface containing a large number of pixels is more robust, and provides great guidance for the subsequent road edge and road centerline predictions. Finally, the results are fused to obtain the segmentation results for the road surface, edge, and, centerlines, respectively. 

\subsubsection{Detailed Convolutional Neural Network Model}
The detailed structure of our road extraction network is shown below in Fig. \ref{fig_roadnet}. As for the backbone network, we use five convolutional stages consisting of thirteen convolutional layers that correspond to the first thirteen convolutional blocks of the VGG-16 \cite{sengupta2019going, simonyan2014very}. One convolutional stage is made up of the convolution, the activation function, and the max pooling. The prominent contributions of our proposed network architecture can be summarized as follows:  
\begin{itemize}
    \item Firstly, we expect the meaningful side output with different scales, and a layer after the fifth pooling yields a too small output plane, which further indicates that the interpolated prediction feature map is too fuzzy to generate a refined result.
    \item  Secondly, the fully connected layers are computationally intensive, which is memory and time consuming. We have proposed a holistically multi-layer fusion network, which comprises a single-stream deep network with multiple side outputs. 
    \item Thirdly, for the convolutional stages one to five, a single-stream deep network derives from VGG-16 is adopted, which is applied to learn multi-scale features and different levels of visual perception. While for side outputs, a $1 \times 1$ and one-channel convolutional layer follows last convolutional layers of each stage. Then, an up-sampling layer is applied to up-sample the feature map. And a sigmoid loss layer is connected to the up-sampling layer in each stage to get the corresponding loss output.
    \item  All the upsampling layers are concatenated by a concatenation layer. Then a one-channel $1 \times 1$ conv layer is applied to fuse the feature maps obtained from each side output. Finally, a loss/sigmoid layer is followed to get the fused loss/output.
\end{itemize}

As shown in Fig. \ref{fig_roadnet}. In stages 1-5, each conv layer is comprised of convolution and scaled exponential linear units (SeLUs) \cite{pedamonti2018comparison}. Here, the convolution is a process with a filter bank to produce a set of feature maps. The spatial pooling is carried out by a max pooling layer, which follows the last convolutional layer of each stage. The feature map size reduction operation is achieved by a stride two block by a max pooling layer with 2$\times$2 pixel filter. In the fusion part, the concatenation layer is a utility layer that concatenates its multiple inputs to one single output. Finally the fused layer gives the semantic prediction after convolutions.

\section{Experiments}
\subsection{Experimental Settings}
\subsubsection{Training Details}
The network proposed is implemented on deep learning framework \textit{Pytorch}. The parameters of the whole convolutional network are initialized with normal distribution. The Bilinear Interpolation is utilized to do up-sampling in the deconvolution process.

 For both the road extraction task and the crack segmentation task, the training epoches is set to 200 and the weight decay is set to 0.9. The learning rate is set initially at 1e-3, and it dropped by five times after every 40 epoches. Our proposed network is efficiently trained without utilizing any pretrained models for the following reasons:
\begin{itemize}
\item Every task aims to distinguish only two classes
(i.e., road and background, edge and non-edge, and centerline and background, cracks and non-cracks), which essentially have fewer classes than the general semantic segmentation.
\item SELU activation function, side output supervision and elaborate loss function facilitate the convergence and improve the accuracy of the proposed network.
\end{itemize}
All the training are done on a single Nvidia GeForce GTX RTX2080Ti GPU.
\subsubsection{Experimental Datasets}
We construct a comprehensive crack segmentation database consisting of 11,298 $450\times450$ cracks images with detailed pixel-level
labels\footnote{\href{https://github.com/KangchengLiu/Crack-Detection-and-Segmentation-Dataset-for-UAV-Inspection}{My Self-Established Database}}. The database contains cracks from various structures, including pavements, bridges, and buildings. In the formulation of domain adaptive crack recognition, we regard the proposed database as the source dataset with label and transfer the model to the pavement-crack and MaWan-crack unlabeled test set for crack recognition with proposed \textit{Crack-DA}. We have also set up our own dataset for the task of road semantic segmentation. We are summarizing the dataset and it will be open-sourced in the future.
\subsubsection{Semi-Supervised Training Settings}
We adopt the semi-supervised settings to conduct the task of semantic segmentation. In our settings, merely 3.5\% of all the images in the training set are labeled and the other images are regarded as the unlabeled data in our semi-supervised learning settings.

\subsection{Road Extraction Results}
We have shown the results with and without our proposed construction loss for the labeled data in Fig. \ref{fig_cons_loss}. It can be demonstrated that without the construction loss, the segmented road pixel ranges are too wide. With our proposed construction loss, we can provide more reliable semantic segmentation results.

Also, we have shown the complete results of the road centerline, edge, and surface predictions in Fig. \ref{edge_results}. It can be seen that our proposed method can provide very accurate segmentation of those three kinds of critical road components. 

Moreover, as shown in Fig. \ref{semi_crack}, to further demonstrate the effectiveness of the proposed approaches in the semi-supervised learning settings, we have also done experiments of the road extraction with merely 3.5\% labeled data. It can be demonstrated that when encounter with limited labeled data, the sup-only method which is merely trained with the supervised optimization function has mistaken and false predictions, while the proposed methods can achieve a very explicit segmentation of various of road structures such as the road surfaces, the road center lines, and the road edges, respectively.

\begin{table*}[htbp!]
\caption{The Comparison of Segmentation Performances Evaluated by Mean Intersection Over Union (MIOU) between Various State-of-the-art Methods with Various Portions of Labeled Data (3.5\%, 5\%, 20\%, 30\%, 50\%, 100\%). The MIOU for the task of crack semantic segmentation is shown on the left of '/', and the MIOU for the road extraction is shown on the right. We have used the threshold (denoted as \textit{Thres}) that can offer the best final segmentation map for each method.}
\label{table_seg}
\begin{center}
\begin{tabular}{c|c|c|c|c|c|c|c|c}
\hline
\toprule
Methods& Backbone Network& \textit{Thres (0-1)} & 3.5\% & 5\% & 20\% & 30\% & 50\% & 100\%\\
\hline
 Ours & \textbf{CR-Seg} &0.49& 53.2/52.1 &56.1/55.2&59.6/58.3 &63.2/62.8 &66.3/65.7 &69.1/ 68.6\\ 

Re-Seg \cite{he2021re} & \textbf{CR-Seg} &0.52&51.2/50.1&53.9/52.8&57.6/56.5&61.7/60.8&63.9/62.7&68.2/68.5\\


EC-Seg \cite{mendel2020semi} &\textbf{CR-Seg} &0.51&49.5/48.6&51.7/50.8&53.9/52.7&56.2/55.2&59.4/58.2&68.7/68.5\\

High-Low-Cons \cite{mittal2019semi} & \textbf{CR-Seg} &0.52&47.2/45.9&49.3/48.2&52.3/51.2&53.5/52.2&57.2/56.3&68.3/67.2\\

CC-Seg \cite{ouali2020semi} & \textbf{CR-Seg} &0.51&44.6/45.2&47.7/47.6&50.7/50.2&52.6/51.0&56.1/55.2&68.6/68.2\\

\bottomrule
\end{tabular}
\end{center}
\end{table*} 




\begin{table*}[t]
\caption{The Detailed Ablation Studies of the Proposed Approach for the Tasks for Crack Semantic Segmentation and Road Extraction under the 5\% labeled circumstance. The MIOU for the task of crack semantic segmentation is shown on the left of '/', and the MIOU for the road extraction is shown on the right.}
\begin{center}
\tiny
\scalebox{1.36}{\begin{tabular}{l|cccccc|c}
\hline
Cases  & Backbone & $L^{m}_{contrast}$ & $L^{m}_{Balance}$& $L_w$ & $L_c$ & Multi-Stage-Fusion & mIOU\% \\
\hline
No. 1 & \checkmark  & \checkmark & \checkmark &  \checkmark & \checkmark & \checkmark &  \textbf{56.1} / \textbf{55.2} \\
No. 2 & \checkmark  & & \checkmark & \checkmark  & \checkmark  & \checkmark &  48.1 / 47.0 \\
No. 3 & \checkmark  &  \checkmark  &  & \checkmark & \checkmark & \checkmark &  51.6 / 50.2 \\
No. 5  & \checkmark  & \checkmark  &  \checkmark &   &\checkmark & \checkmark &  55.6 / 54.8\\
No. 6 & \checkmark & \checkmark & \checkmark  & \checkmark &  &\checkmark &  53.7 / 52.6\\
No. 7 & \checkmark  & \checkmark & \checkmark &  \checkmark & \checkmark &  & 53.2 / 52.2 \\
\hline
\end{tabular}}
\label{table_ablation}
\end{center}
\setlength{\abovecaptionskip}{-0.88cm}
\setlength{\belowcaptionskip}{-0.88cm}
\end{table*}
\subsection{Crack Semantic Segmentation Results}
Also, as shown in Fig. \ref{semi_crack}, to further demonstrate the effectiveness of the proposed approaches in the semi-supervised learning settings, we have also done experiments of the road extraction with merely 3.5\% labeled data for the task of crack semantic segmentation. It can be demonstrated that when encounter with limited labeled data, the sup-only method which is merely trained with the supervised optimization function has mistaken, false, and vague predictions, while the proposed methods can achieve a very explicit segmentation of various of various crack patterns under complex background objects such as shadows and non-crack noises.
\subsection{Quantitative Comparisons of the Proposed Methods with State-of-the-arts Semi-supervised Semantic Segmentation Methods}
As shown in Tabel \ref{table_seg}, we have compared our proposed method with those SOTAs semi-supervised semantic segmentation counterparts, including Re-Seg \cite{he2021re}, EC-Seg \cite{mendel2020semi}, High-Low-Cons \cite{mittal2019semi}, and CC-Seg \cite{ouali2020semi}. We have used the threshold (denoted as \textit{Thres}) that can offer the best final segmentation map for each method. It can be demonstrated that our proposed approach achieves significant improvement based on the previous SOTAs methods. For example, in the circumstances of merely 3.5\% labeled data, our proposed method outperforms previous best method  Re-Seg \cite{he2021re} by a large margin of 2.0\% for the task of crack semantic segmentation, and by a margin of 1.9\% for the task of road extraction. It can be also seen that our proposed approach has great performance in the circumstances that the training data is extremely limited (e.g. the 3.5\% and the 5\% labeled case). And our proposed method has nearly equal performance with those semi-supervised counterparts when the is fully labeled (e.g. the labeling percentage is 100\%). It demonstrates that our proposed method has superior performance when the labeled data is extremely limited.
\vspace{-2mm}
\subsection{Ablation Studies}
As shown in Table \ref{table_ablation}, we have also done very detailed ablation studies for various of proposed network modules. We have ablated in all settings including the following settings:
\begin{enumerate}
    \item The full proposed network framework (Case No. 1);
    \item Discarding our proposed confidence-level based contrastive learning optimization function (Case No. 2);
    \item Discarding our proposed data balance optimization function (Case No. 3);
    \item Discarding our proposed weight decay optimization function (Case No. 4);
    \item Discarding our proposed construction optimization function (Case No. 5). 
     \item Discarding our proposed multi-stage fusion strategy in the optimization function (Case No. 6).
\end{enumerate}

It can be demonstrated that when discarding any proposed network modules for the global optimization, the network performance will drop. And the proposed confidence-level based contrastive learning strategy has great influence on the network performance. Discarding the contrastive loss will results in a very significant drop in the final network performance. The ablation study has further demonstrated the effectiveness of each module in our proposed framework for the tasks of both crack semantic segmentation and the road extraction, respectively. 

\section{Conclusions}
In this work, to overcome the data hungry challenge, we have proposed a semi-supervised contrastive learning framework for the task of class-imbalanced semantic segmentation. First and foremost, to make the model operates in a semi-supervised manner, we proposed the confidence-level based contrastive learning to achieve instance discrimination in an explicit manner, and make the low-confidence low-quality features align with the high-confidence counterparts. Moreover, to tackle the problem of class imbalance, we proposed the data imbalance loss to replace the traditional cross entropy loss in pixel-level semantic segmentation. Extensive experiments on the real industrial crack segmentation and the road segmentation demonstrate the superior effectiveness of the proposed framework. Our proposed method can provide satisfactory segmentation results with merely 3.5\% labels. 

\addtolength{\textheight}{0cm}   





\bibliographystyle{IEEEtran}
\bibliography{references}

\end{document}